\documentclass{article}

    \PassOptionsToPackage{numbers, sort&compress}{natbib}
\usepackage[preprint]{neurips_2019}

\usepackage[utf8]{inputenc} % allow utf-8 input
\usepackage[T1]{fontenc}    % use 8-bit T1 fonts
\usepackage[colorlinks,bookmarks=false]{hyperref}       % hyperlinks
\usepackage{url}            % simple URL typesetting
\usepackage{booktabs}       % professional-quality tables
\usepackage{amsfonts}       % blackboard math symbols
\usepackage{nicefrac}       % compact symbols for 1/2, etc.
\usepackage{microtype}      % microtypography

\usepackage{times}
\usepackage{epsfig}
\usepackage{graphicx}
\usepackage{amsmath}
\usepackage{amssymb}
\usepackage{pifont}
\usepackage{enumitem}
\usepackage{svg}
\usepackage[export]{adjustbox}
\usepackage{algorithm}
\usepackage{multirow}
\usepackage{mmstyles}
\usepackage{breqn}
\usepackage{csquotes}
\usepackage[export]{adjustbox}
\usepackage{algpseudocode}

\title{Biased Estimates of Advantages over Path Ensembles}

\author{%
  Lanxin Lei\(^*\)\hspace{3em} Zhizhong Li\thanks{The two authors contributed equally.}\hspace{3em} Dahua Lin\\
  \\
  Department of Information Engineering, The Chinese University of Hong Kong\\
  Shatin, Hong Kong, China \\
  \texttt{leilansen@gmail.com, \{lz015,dhlin\}@ie.cuhk.edu.hk} \\
}

\begin{document}

\maketitle

% !TEX root = main.tex

\begin{abstract}
The estimation of advantage is crucial for a number of reinforcement learning algorithms, as it directly influences the choices of future paths. In this work, we propose a family of estimates based on the order statistics over the path ensemble, which allows one to flexibly drive the learning process, towards or against risks. On top of this formulation, we systematically study the impacts of different methods for estimating advantages. Our findings reveal that biased estimates, when chosen appropriately, can result in significant benefits. In particular, for the environments with sparse rewards, optimistic estimates would lead to more efficient exploration of the policy space; while for those where individual actions can have critical impacts, conservative estimates are preferable. On various benchmarks, including MuJoCo continuous control, Terrain locomotion, Atari games, and sparse-reward environments, the proposed biased estimation schemes consistently demonstrate improvement over mainstream methods, not only accelerating the learning process but also obtaining substantial performance gains.
% In the estimation of advantage, if the trajectory is longer, then the estimation would be more accurate and effective, but it suffers from the averaging effect, i.e., the effective actions are encumbered by the imperfect later actions.
% We can take use of the shorter estimations to form an ensemble,
% to reduce the averaging effect.
% The proposed method can be seen as an exploration bias,
% because it exaggerates the estimation of advantages:
% if we think it is large, then we take the largest estimation in the ensemble, and if we think it is small, we take the smallest estimation.
% In actor-critic family of algorithms, the method basically encourages actions that have the `potential' of being good,
% and discourages actions that `seems' to be bad.
% The method is applicable for various current DL algorithms (such as PPO, A2C, etc.) and effective for various RL environments.
% We verified the effectiveness of the method in 8 MuJoCo environments using PPO\@.
% For sparse environments,
% the averaging effect would be more obvious.
% So based on EX2, an dedicated method for sparse-reward environments,
% we verified that the proposed algorithm not only accelerates the training speed, but also improves the performance.
\end{abstract}

% !TEX root = main.tex

\section{Introduction}\label{sec:intro}

The research on deep reinforcement learning is gaining momentum in recent years.
A number of learning methods, such as
Proximal Policy Optimization algorithm (PPO)~\cite{schulman2017proximal},
Trust-Region Policy Optimization algorithm (TRPO)~\cite{schulman2015trust}, and
Advantage Actor-Critic algorithm (A2C)~\cite{mnih2016asynchronous},
have been developed.
These methods and their variants have achieved great success in challenging problems,
\eg~continuous control~\cite{schulman2015high},
locomotion~\cite{heess2017emergence},
and video games~\cite{vinyals2017starcraft}.

The core of all these methods is the estimation of the \emph{advantage}
\(A(s_t, a_t)\), \ie~the gain in the expected cumulative reward relative to the state value \(V(s_t)\)
if a certain action \(a_t\) is taken.
A common practice is to use the \(k\)-step estimate over the sampled trajectories.
This way has been widely used in actor-critic algorithms such as the A2C.
%  take the average of the
%
Schulman~\etal presents a generalization called
Generalized Advantage Estimator (GAE)~\cite{schulman2015high}, which
combines different \(k\)-step estimators with exponentially decayed
weights, thus reducing the variance of policy gradients.
For convenience, we refer to the set of \(k\)-step estimators along a trajectory with different \(k\) values as the \emph{path ensemble}.

Whereas taking the linear combination over the path ensemble as in GAE
is reasonable from a theoretical view, it is not necessarily an effective
strategy in practice, especially in challenging environments.
For example, in the environments with sparse rewards, it might be advisable to
take a more optimistic view, actively exploring those policies that show
potentials without being discouraged by a few failed paths; while
in the environments that are fragile, \eg~those where a wrong choice of action
can lead to catastrophic consequences down the path, a more conservative, risk-averse
stance might be preferable.
Inspired by the observation that different environments may require different
strategies in exploring the optimal policies, we propose a new family of
\emph{biased} estimators of the advantage based on order statistics.
The proposed method takes the maximum, minimum, or other generalized order statistics
over the path ensemble as the overall estimates of the advantage,
depending on the characteristics of the environment.

We systematically studied different estimation schemes on various environments,
including
the sparse-reward environments~\cite{NIPS2017_6851},
the Terrain RL Simulator~\cite{Berseth2018TerrainRS} for locomotion,
the Atari games~\cite{bellemare13arcade},
and the set of MuJoCo environments for continuous control~\cite{gym}.
Our study shows that the proposed estimation strategies, when chosen
appropriately, can substantially outperform the standard way.
In particular, optimistic estimation, \ie~taking the maximum,
greatly improves the learning efficiency
in the environments with sparse rewards.
On the other hand, pessimistic estimation, \ie~taking the minimum,
can effectively stabilize the learning by avoiding risky actions.

% However,
% using only the average over the path ensemble is a waste of resource:
% the distribution of these estimators also conveys useful information.
% We focus on the nonlinear combination over the path ensemble,
% such as the \(\max\), the \(\min\), and the \(\maxabs\) rules.
% Their respective effects can be connected to the optimistic, pessimistic and exaggerated policies.
% For example,
% the \(\max\) over the path ensemble give a trajectory more tolerance of bad decisions.
% Suppose in the middle of the trajectory,
% a bad action was accidentally taken,
% then its influence might be suppressed by the \(\max\) operator.
% This leads to an optimistic estimation of the advantage:
% a large value is returned if one of the estimation is large.
% As will be demonstrated in the experiments,
% the optimistic estimation greatly improves the sample efficiency in sparse-reward environments.

% explain intuitively how nonlinear path ensemble relates to risks
% there are also something more to consider

It is noteworthy that
the biased estimators proposed in this work differ essentially from
the recent two lines of work on
robustness~\cite{smirnova2019distributionally,delage2007percentile} and
risk-sensitivity~\cite{tamar2012policy,chow2014algorithms}.
Robust MDP optimizes for the worst case when there exist uncertainties
in the parameters; risk-sensitive MDP optimizes the value of a risk measure;
while our method does not modify the optimization objective itself.
Instead, it controls
the bias of the policy gradient towards policies of different styles through the alternative ways
of estimating the advantage.

% Ours instead conditioned on different trajectories.

% what else can they be used to

%   risk-sensitive, robust, distributional

% the difficulty in estimation

%   need samples for the estimation starting from the same state
%   difficult to acquire
%   unless amortized

% what is our idea

%   for a path, we have an ensemble of estimations
%   though they are not independent, but they are mostly wasted
%   we want to take full usage of them

% We conducted extensive experiments on different types of environments,
% including environments with sparse reward that appeared in Exploration with Exemplar models (EX2)~\cite{NIPS2017_6851},
% the Terrain RL Simulator~\cite{Berseth2018TerrainRS} for locomotion,
% the Atari games~\cite{bellemare13arcade},
% and the set of MuJoCo environments for continuous control~\cite{gym}.
% % and the StarCraft II minigame environment~\cite{vinyals2017starcraft} with delayed and sparse reward.
% The nonlinear path ensemble methods demonstrate the superiority of performance in suitable environments.

%%

% the bias-variance between

%%

% contribution

% experimental results

% summary

\section{Related Work}\label{sec:related}

\paragraph{Estimation of Advantage and Action-Value}
Since the advantage \(A(s,a)\) is simply the difference between the action-value function \(Q(s,a)\) and the value function \(V(s)\),
estimators for \(Q\) can be adapted to estimate \(A\).
% In tabular cases,
% the \(n\)-step TD methods,
The \(k\)-step advantage estimation is derived from the \(k\)-step return,
which is used in the \(k\)-step TD or the \(k\)-step Sarsa method~\cite{sutton2011reinforcement}.
Similarly, the generalized advantage estimator~\cite{schulman2015high} is analogous to the \(\lambda\)-return~\cite{watkins1989learning} in the TD(\(\lambda\)) method~\cite{seijen2014true}. % chktex 36
% In the \(n\)-step Sarsa method,
%  is used to estimate the action-value function \(Q(s,a)\).
% The estimation of advantage is closely related to the estimation of the \(Q\) and value
% Compared to the two extreme cases of \(1\)-step TD or \(\infty\)-step TD,
% the \(n\)-step estimations are often
% As we have mentioned earlier,
They are all linear combinations over the path ensemble.
A nonlinear combination scheme is seen in the work of Twin Delayed DDPG (TD3)~\cite{fujimoto2018addressing},
which uses the \(\min\) of two critics to estimate \(Q\).
However, it aims to mitigate the overestimation problem in deep Q-learning,
and the objects be combined are the network outputs instead of the estimators based on the cumulative rewards.
Another nonlinear combination is the Positive Temporal Difference (PTD)~\cite{van2007reinforcement,vanHasselt2012},
where the advantage is set as \(1\) when the \(1\)-step estimate of advantage is positive,
and is set to \(0\) otherwise.
The benefit of limiting policy updates only toward those actions that have positive advantages is increasing the stability of learning.
% A notable difference of the order statistics over ensembles to the linear combinations is that
% the former dependents on the specific values of the estimators,
% while the latter uses fixed combination coefficients for the estimators.
% We use the nonlinear combination of path ensembles to
% In value-based learning method,
% the estimation of \(Q\)-function is important.
% \(1\)-step estimation,
% \(n\)-step estimation.
% Generalized Advantage~\cite{schulman2015high}.
% TD3

\paragraph{Robustness and Risk-Sensitivity in RL}
In the assumption of robust MDPs~\cite{nilim2004robustness,xu2007robustness,delage2010percentile,mannor2012lightning,smirnova2019distributionally},
the parameters of the problem lie in an uncertainty set,
and the target is to find a solution that performs best under the worst cases.
On the other hand, risk-sensitive MDPs consider the uncertainty of the rewards.
The objective is to minimize the risk-measure,
which is defined by the exponential utility~\cite{howard1972risk},
the Conditional Value-at-Risk (CVaR)~\cite{chow2014algorithms,prashanth2014policy,chow2015risk,tamar2015optimizing},
the percentile~\cite{delage2007percentile} or the variance~\cite{tamar2012policy,prashanth2013actor} of the cumulative rewards.
% Risk-sensitive MDPs are close related to robust MDPs~\cite{osogami2012robustness},
% Risk-Sensitive and Robust Decision-Making: a CVaR Optimization Approach (NIPS 2015)~\cite{chow2015risk},
% Optimizing the CVaR via Sampling (AAAI 2015)~\cite{tamar2015optimizing},
% Algorithms for CVaR Optimization in MDPs (NIPS 2014)~\cite{chow2014algorithms},
% Actor-Critic Algorithms for Risk-Sensitive MDPs (NIPS 2013)~\cite{prashanth2013actor},
% Percentile Optimization for Markov Decision Processes with Parameter Uncertainty (2009)~\cite{delage2010percentile},
% Robustness and risk-sensitivity in Markov decision processes (NIPS 2012)~\cite{osogami2012robustness},
% Policy gradients for CVaR-constrained MDPs (2014)~\cite{prashanth2014policy},
% Policy Gradients with Variance Related Risk Criteria~(ICML 2012)~\cite{tamar2012policy},
% Percentile Optimization in Uncertain Markov Decision Processes with Application to Efficient Exploration (ICML 2007)~\cite{delage2007percentile},
% Distributionally Robust Reinforcement Learning (ICML 2018)~\cite{smirnova2019distributionally}
% Opportunistic Learning: Budgeted Cost-Sensitive Learning from Data Streams (ICLR 2019)~\cite{kachuee2019opportunistic},
% The statistics over the set of path ensembles is not same as directly estimating them.
The statistics over the path ensemble is not the same as those statistics for advantage.
% For example, \(\Ebb\big[\max\{\hat A^{(k)}_t\}_{k=1}^n\big]\) is clearly less than \(\max \hat A^n_t\).
However, using a suitable nonlinear combination of the numbers in the path ensemble,
the exploration strategy in our algorithms can bias the learning toward risk-averse or risk-seeking policies.
Unlike the robust and risk-sensitive approaches,
which typically introduce involved estimations,
our approach is straightforward to implement and requires minimal changes to incorporate the idea into existing algorithms such as A2C, PPO, TRPO and their variants.

\paragraph{Distributional RL}
The value distribution is the object of study in distributional RL\@.
Since the full distribution of return is difficult to estimate,
researchers have adopted nonparametric methods~\cite{morimura2010nonparametric} or used simple distributions~\cite{bellemare2017distributional,dabney2018implicit,barth2018distributed} to approximate it. %, which is difficult to estimate.
% The distribution can be estimated
% In distributional RL, the full distribution of the
% Nonparametric Return Distribution Approximation for Reinforcement Learning (ICML 2010)~\cite{morimura2010nonparametric},
% A Distributional Perspective on Reinforcement Learning (ICML 2017)~\cite{bellemare2017distributional},
% Distributional Reinforcement Learning with Quantile Regression (AAAI 2018)~\cite{dabney2018distributional},
% Implicit Quantile Networks for Distributional Reinforcement Learning (ICML 2018)~\cite{dabney2018implicit},
% Distributed Distributional Deterministic Policy Gradients (ICLR 2018)~\cite{barth2018distributed}.
The RL algorithms can be formulated with any criterion based on the distribution of return,
such as the aforementioned risk-sensitivity measures.
Again, the distribution formed by elements in the path ensemble does not represent the full distribution of return,
but some joint distribution that we can exploit.

% \paragraph{Reward Shaping}
% Reward re-distribution

\section{Method}\label{sec:method}

In this section,
we first review the preliminary knowledge on RL and setup notations.
Then we introduce the central concept of this article: the biased estimators over path ensembles.
Specifically, we focus on the family of order statistics, \eg, the maximum value.
% In statistics, the \(k\)-th order statistics is equal the \(k\)-th smallest value of a statistical sample.
% For example, is one special case of order statistics.
Next,
we give an illustrative study on the \(\max\) statistics over the path ensemble,
where we show that the induced biased estimator of action-values would influence the optimization process of RL algorithms.
With this specific example in mind,
we give a practical algorithm to incorporate the general biased estimators into existing RL algorithms.
Lastly, we discuss the influence of different biased estimators on the learning process.

\subsection{Preliminary}\label{sec:pre}

We consider the standard formulation of RL,
which is typically modeled as an MDP \(\left( \cS, \cA, \cT, \gamma, R \right)\),
where \(\cS\) is the state space,
\(\cA\) is the action space,
\(\cT=\left\{P_{sa}(\cdot)\mid s\in \cS, a\in \cA\right\}\) is the transition probabilities,
\(\gamma\in (0,1]\) is the discount factor, % chktex 9
and \(R\) is the reward function.
At timestep \(t\),
the agent in state \(s_t\) interacts with the environment by choosing an action \(a_t\sim \pi(s_t)\) following the policy \(\pi\),
and receives a reward \(R_t\) from the environment.
% Here \(\cP(\cA)\) is the set of probability measures on the action space \(\cA\).
The environment then transits to the next state \(s_{t+1}\).
The discounted return is defined as \(G_t := \sum_{i=t}^{T-1} \gamma^{i-t} R_{i}\),
and the goal of RL is to maximize the expected return \(J = \Ebb_{s_0\sim S_0}\left[G_0\mid s_0\right]\),
where \(S_0\) is the distribution of initial states.
The action-value function \(Q^{\pi}(s_t,a_t) := \Ebb\left[G_t \mid s_t, a_t \right]\) under a policy \(\pi\) is the expected return of taking action \(a_t\sim\pi(s_t)\) in state \(s_t\).
% \(Q^*\) for the optimal policy \(\pi^*\) satisfies the optimal Bellman equation,
% \begin{equation}
%     {Q^{*}(s,a)}
%     = \Ebb_{s'\sim P_{sa}}\left[ R(s,a,s') + \gamma \max\limits_{a'\in \cA}Q^{*} (s', a')\right].
% \end{equation}
In policy-based model-free deep RL algorithms,
policy \(\pi_\theta\) with parameters \(\theta\) is optimized via gradient ascent.
An estimation of the gradient is the policy gradient
\begin{equation}\label{eq:grad}
    \nabla_{\theta} J(\theta) = \Ebb_{\tau\sim\pi_\theta(\tau)}
    \bigg[\sum_{t=0}^{T} \frac{\nabla_\theta\pi_\theta(a_t \mid s_t)}{\pi_\theta(a_t \mid s_t)} A(s_t,a_t)\bigg],
\end{equation}
where \(\tau=\{s_0,a_0,s_1,a_1,\ldots,s_T,a_T\}\) is a trajectory following the policy.
\(A(s_t,a_t):=Q(s_t, a_t)-V(s_t)\) is the advantage function.
% Given a trajectory,
%  \( (s_t, a_t, r_t, s_{t+1}, a_{t+1}, r_{t+1}, \dots, s_T) \),
The \(k\)-step estimation of the advantage function is given by
\begin{equation}\label{eq:ak_def}
    \hat A_t^{(k)} := \sum_{i=0}^{k-1} \left(\gamma^i r_{t+i}\right) + \gamma^k V(s_{t+k}) - V(s_{t}), \quad k=1,2,\dots,T-t.
\end{equation}
% generalized advantage
% \begin{equation}
    %     \GAE(\gamma, \lambda) := (1 - \lambda) \left( \hat A_t^{(1)} + \lambda \hat A_t^{(2)} + \lambda^2 \hat A_t^{(3)} + \cdots \right).
% \end{equation}
The generalized advantage estimator \(\GAE(\gamma, \lambda)\) is an exponentially-weighted average of different \(k\)-step estimators,
where
\(\GAE(\gamma, \lambda) := (1 - \lambda) \big( \hat A_t^{(1)} + \lambda \hat A_t^{(2)} + \lambda^2 \hat A_t^{(3)} + \cdots \big)\).

\subsection{Biased Estimators over the Path Ensemble}

Given a trajectory,
we define the set of \(k\)-step estimators \(\cE:=\big\{ \hat A_t^{(1)}, \hat A_t^{(2)}, \dots \big\}\) as the \emph{path ensemble} for the pair \((s_t, a_t)\).
The GAE is simply a linear combination of elements in this path ensemble.
% In this article, w
We set out to explore the effects of \emph{nonlinear} combinations.
Specifically,
we are interested in the family of order statistics,
where the \(k\)-th order statistics equals to the \(k\)-th smallest value in the path ensemble.
The following special cases of the (generalized) order statistics will be studied in detail.
\begin{enumerate}
    \item The \(\max\) statistics and the \(\min\) statistics,
    \begin{equation}
        \hat A^{\text{max}}_t := \max_i\Big\{ \hat A_t^{(i)} \Big\}, \quad \hat A^{\text{min}}_t := \min_i\Big\{ \hat A_t^{(i)} \Big\}.
    \end{equation}
    \item The \(\maxabs\) statistics,
    \begin{equation}
        \hat A^{\text{max-abs}}_t := \argmax_{A\in\big\{ \hat A_t^{(i)} \big\}}\; \lvert A \rvert. %, \quad
        % \hat A^{\text{min-abs}}_t := \argmin_{A\in\big\{ \hat A_t^{(i)} \big\}}\; \lvert A \rvert.
    \end{equation}
    It is a generalization of order statistics where the order is counted according to the absolute value of each element instead of the element itself.
    \item The general order statistics \(\hat A^{\text{order-}d}_t\), which is the \(d\)-th smallest element in the set \(\cE\).
    % \begin{equation}
    %      := \order \Big(d, \Big\{ \hat A_t^{(i)} \Big\} \Big).
    % \end{equation}
\end{enumerate}

% \begin{equation}
%     \sum_{t=0}^{k-1} \gamma^t R_{t} + \gamma^k V(s_{k})-V(s_{t}),
% \end{equation}
% where \( V(s_t)=\Ebb_{a_t}\left[Q(s_t,a_t)\right]\) is the value function and \(k\) is bounded by \(T\).

% value-based, policy-based, actor-critic

\subsection{An Illustrative Study on the \(\max\) Statistics}\label{sec:q}

\begin{figure}[t]
	\centering
	\includegraphics[width=0.99\linewidth]{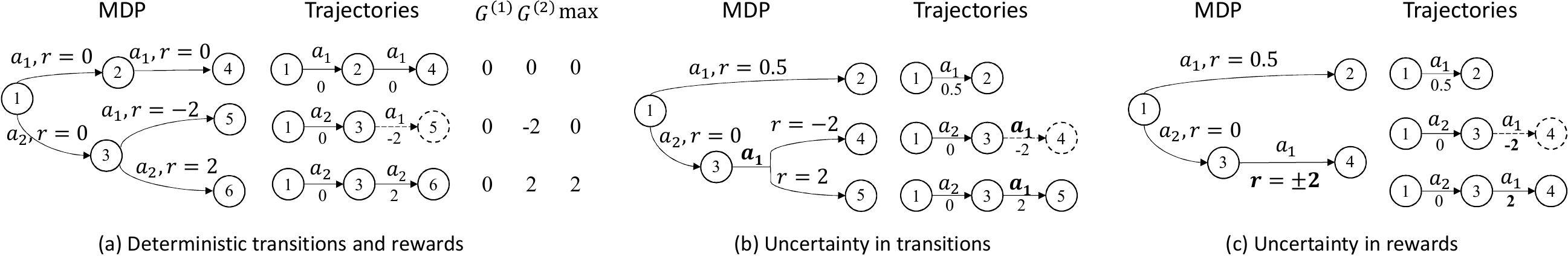}
	\caption{
		Toy examples using the \(\max\) path ensemble.
		(a) When calculating return of the second trajectory,
		the \(\max\) chooses \(G^{(1)}\), so that the bad action \(a_1\) at state \(s_3\) is blacklisted.
		This leads to an optimistic view when evaluating the long-term returns of actions.
		The dashed circle means that it is ignored in the computation of return.
		(b) and (c) demonstrate the over-estimation problem of the \(\max\) statistics when uncertainty exists in transitions or rewards.
	}\label{fig:toy}
\end{figure}

Before delving into the discussion of estimating advantages,
% we first look at a concrete example of how the \(\max\) statistics induces good actions with high probability.
we first look at a concrete example of how the \(\max\) statistics of \(Q\)-values affect the learning process of
% %  of the policy iteration algorithm.
% The \(\max\) statistics is used to discover actions that have the potential to be good.
% To illustrate how it works, we give a toy example using
the policy iteration algorithm~\cite[Section~4.3]{sutton2011reinforcement}.
In Figure~\ref{fig:toy}~(a),
an MDP with 6 states is drawn.
For simplicity, we assume that both the state transitions and the rewards are deterministic,
and the discount factor is \(1\).
In this example, only the two actions in state \(s_3\) can get rewards,
where \(R(s_3,a_1, s_5)=-2\) and \(R(s_3,a_2, s_6)=2\).
At the initialization step of the policy iteration algorithm,
the policy is initialized as the uniform random policy~\(\pi_1\).
In the first step of the policy iteration,
we evaluate the \(Q\) function and value function \(V\) of this random policy \(\pi_1\).
Under policy \(\pi_1\), all possible trajectories are also shown in Figure~\ref{fig:toy}~(a).
We compute that \(Q^{\pi_1}(s_1, a_1)=0\), and \(Q^{\pi_1}(s_1, a_2)=0\).
%  and \(Q^{\pi_1}(s,a)=0\) for all other state \(s\) and action \(a\).
So in the next step,
the greedy policy \(\pi_2\) for state \(s_1\) is still at random.
%  because \(Q^{\pi_1}(s_1,a_1)=Q^{\pi_1}(s_1,a_2)=0\).
However,
% based on the exact value function,
we can have a biased estimation \(\hat Q^{\text{max}}\) using the max statistics,
% Specially,
\begin{equation}
\hat Q^{\text{max}}(s_1, a_1)=\Ebb\left[\max_i\big\{ R^{(i)} \big\} \mid s_1, a_1\right] = 0,\quad
\hat Q^{\text{max}}(s_1, a_2)=\Ebb\left[\max_i\big\{ R^{(i)} \big\} \mid s_1, a_2\right] = 1.
\end{equation}
Using \(\hat Q^{\text{max}}\),
the greedy policy at state \(s_1\) is \(\argmax_a \hat Q^{\text{max}}(s_1, a) = a_2\).
Actually, the algorithm with \(\max\) statistics finds the optimal policy within only \(1\) step in this example.
In comparison, the original policy iteration needs \(2\) steps to reach optimal.

% In the above example,
We see that the \(\max\) statistics over the path ensembles converges faster because it discovers the potentially good action \(a_2\) at state \(s_1\) earlier than using the standard way.
As the computation of the second trajectory in Figure~\ref{fig:toy}~(a) shows,
the \(\max\) statistics blacklists the bad action \(a_1\) in state \(s_3\).
In a trajectory,
when an action is so bad that the later actions cannot compensate for the caused loss,
it will be replaced by the ``average actions''.
This good-action discovering ability is suitable for sparse-reward environments.
In the early exploration stage of the training in those environments,
the agent barely receives positive reward signals.
Many trajectories might be only partially good,
namely,
it happens to act well at some time,
but then go to bad states later due to wrong actions.
The \(\max\) statistics can highlight actions that are possible to get high rewards in any middle step of the trajectory,
even if the overall return which considers till the end of the trajectory is very low.
By effectively discovering good actions,
this method is expected to improve the sampling efficiency.

Note that when uncertainty presents in either the state transitions or the rewards,
applying the \(\max\) statistics in policy iteration may fail to improve policy.
We give two examples when the estimation is overly optimistic.
The first example, which is shown in Figure~\ref{fig:toy}~(b),
is caused by the uncertainty in state transitions.
In this example,
we have \(\hat Q^{\text{max}}(s_1,a_1)=0.5\) and \(\hat Q^{\text{max}}(s_1,a_2)=1\),
which implies that action \(a_2\) should be chosen in state \(s_1\).
However, the optimal strategy is to select \(a_1\).
This is caused by the the \(\max\) statistics' ignorance on the coupling risks.
When taking an action, both good and bad may happen.
The bad cases is ignored in the estimation.
%  and make it irrelevant to the value function estimation.
However, we cannot avoid bad next-state when selecting that action.
The second example, shown in Figure~\ref{fig:toy}~(c), is a case when the reward owns randomness.
The problem is that the maximum operation over the ensemble causes overestimation,
a symptom that also troubles the \(Q\)-learning algorithm as discussed in TD3~\cite{fujimoto2018addressing}.

\subsection{Incorporating Biased Advantage Estimators into RL Algorithms}\label{sec:adv}

We have seen that the \(\max\) statistics is good at discovering potentially good actions,
but purely using this biased estimation in policy iteration may lead to sub-optimal solutions.
%
% First, we want to exploit the benefit of the biased estimation;
% and second, we want the algorithm to have better performance.
To overcome this problem,
% while still take advantage of the benefit of the biased estimation,
the new estimation is used as an \emph{exploration strategy}.
Namely,
% we actually performs a mixed strategy,
the biased estimation is used with probability \(\rho\),
and the original estimation is used otherwise.
The hyper-parameter \(\rho\) is named as the \emph{bias ratio}.
% instead of completely replacing the role of the original estimation.
This discussion also applies to the estimation of advantages.

% Now we turn to the discussion of advantage estimation.
% Advantage is used in the actor-critic family of algorithms.
The foundational position of advantage in actor-critic algorithms is attributed by the policy gradient theorem in Equation~\eqref{eq:grad}.
% There are different forms of the theorem.
% In A2C, the original form is used.
% In TRPO, modification.
% In PPO, more modifications.
A sample \((s_t, a_t)\) makes the policy network \(\pi_\theta\) adjust its parameters according to the estimation of advantage \(\hat A_t(s_t, a_t)\).
The probability of \(\pi_\theta(s_t)=a_t\) is increased when the advantage is positive,
and decreased when the advantage is negative.
Thus the value of advantage is critical to the optimization direction of the algorithm.
When the estimation of certain advantages \(\hat A_t(s_t, a_t)\) is manipulated,
the optimization process can be biased toward the desired style of policies.
% The following observation is the tool our approach relies on.
% An important observation is that,
% for the policy-gradient based algorithms,
For example,
when the estimation of advantage \(\hat A_t(s_t, a_t)\) is manipulated to be larger,
the learned policy is then biased toward the action \(a_t\) at state \(s_t\).

% \begin{obs}
%     In the policy gradient family of algorithms,
%     if we manipulate the estimation of advantage \(\hat A_t(s_t, a_t)\) to be larger,
%     then the learned policy would biased toward action \(a_t\) in state \(s_t\).
%     % and \emph{vice versa}.
%     % For PPO, analyze policy gradient, how does the biased exploration affects the policy.
% \end{obs}

% Now we consider how to create the desired biased estimation using the estimators in the path ensemble.
% First we look at the inter-relationships of those estimators.
% Some notable observations of advantage-estimators in the path ensemble.
% \begin{obs}
%     For the \(k\)-step estimator of advantage, the bias generally decreases, and variance increases as \(k\) increases.
% \end{obs}
Note that the elements in the path ensemble are inter-related.
Roughly speaking, the difference between the \(i\)-step advantage estimator \(\hat A_t^{(i)}\) and the \(j\)-step estimator \(\hat A_t^{(j)}\) in a path ensemble is discounted exponentially with the minimum step index \(\min(i,j)\) of the two.
In fact, let \(j>i\),
then
\begin{equation}
    \hat A_t^{(j)} - \hat A_t^{(i)} = \gamma^i\left( \sum_{s=0}^{j-i-1} \left(\gamma^{s} r_{t+s+i}\right) + \gamma^{j-i} V(s_{t+j}) -  V(s_{t+i})\right).
\end{equation}
The sum in the big bracket can be assumed to be bounded.
Based on this observation,
the values of the estimation \(\hat A_t^{(k)}\) would not change too much when \(k\) becomes large.
In practice we can use a subset of the full ensemble %without loss of much information.
%  near estimators have strong correlation.
to reduces the computational cost.
%  we use a subset of all possible \(k\)-step estimators to form the ensemble.
% In practice, we use part of the estimators to form the ensemble.
For example, we used an ensemble with only four elements
\(\cE=\{\hat A_t^{(1)},\hat A_t^{(16)},\hat A_t^{(64)},\hat A_t^{(2048)}\}\) in the MuJoCo physics simulator experiments.
% where the last one \(\hat A_t^{(2048)}\) is typically uses the longest possible trajectory in one batch.
% Another benefit of not using the full set of ensemble is that,
% The order statistics get stabilized when it is taken over a large enough set.
% For example, the distribution of the maximum of \(n\) iid uniform random variables is approximately the constant \(1\) when \(n\) goes to infinity.
% This effect may leads the order statistics lose information if the set is too large.
% % The comparison of using different set of path lengths can be found in experiments,
% This is verified in the experiment section,
% where we observed performance degradation of \(\max\), \(\min\) and \(\maxabs\) statistics from ensemble of size \(4\) to size \(12\).

% % Explains why we irregular sample of numbers.
% Explain why they can approximate optimistic, risk-averse policies.
% They regularize the actual behavior.

% In the estimation of advantage, uses nonlinear function, with the cost of introducing bias.
% To be specific,
% we focus on three statistics: \(\max\), \(\min\) and \(\maxabs\).

Summarizing the above observations,
our algorithm is designed as follows.
%
% \vspace{-0.1cm}
\begin{algorithm}[H]
    \caption{Biased Estimates of Advantages over Path Ensembles}\label{TEST}
    \begin{algorithmic}[1]
        \State Parameters: the biased estimation statistics  \(\hat A^{\text{biased}}\), the ensemble set \(\cE\) and the bias ratio \(\rho\).
        \For{{each iteration}}
        	\State {Collect trajectories}.
        	\State {Compute the normal advantage \(\hat A\) and the biased estimation \(\hat A^{\text{biased}}\).}
        	\State {Update the policy network according to PPO, A2C, or TRPO, in which \(\hat A^{\text{biased}}\) is used with probability \(\rho\), and \( \hat A\) is used otherwise.}
        	\State {Update the critic network using the normal estimation of \(Q\) value, which equals to \(V+\hat A\).}
        \EndFor
    \end{algorithmic}
\end{algorithm}
% \vspace{-0.1cm}
%
% \begin{algorithm}[H]
%     \caption{Order Statistics over Path Ensemble}
%     \begin{algorithmic}
%         \STATE Parameters: the biased estimation statistics \(\hat A^{\text{biased}}\), the ensemble index set and the bias ratio \(\rho\).
%         \STATE For each iteration
%         \STATE collect trajectories.
%         \STATE compute the normal advantage estimation \(\hat A\) and the biased estimation \(\hat A^{\text{biased}}\)
%         \STATE with probability \(\rho\), update the policy using \(\hat A^{\text{biased}}\), and otherwise use \(\hat A\).
%         \STATE always update the critic using the normal estimation of \(Q\)-value \(V+\hat A\).
%     \end{algorithmic}
% \end{algorithm}
%
Note that the critic is not updated with the biased estimation.
This keeps the estimation of values from being affected by the biased estimation.
Our method can be easily plugged into any actor-critic algorithms with the minimal computational cost.

% The implementation is simple and the  is minimal.

\subsubsection{Discussion}

% The path ensemble focus on the behavior along trajectories.
% The estimations is different from estimations across trajectories.
% This is the fundamental difference between our approach and the robust/risk-sensitive/distributional MDPs.
% Another way of estimating statistics: not very exact, but useful and easy to implement.
% For example, the \(\min\) statistics implies the risk-averse bias.
% This risk-averse is explained as,
% along any trajectory, we avoid bad actions happens.
%
The biased estimator over path ensemble focuses on the behavior along trajectories,
which is the fundamental difference between our approach and the robust/risk-sensitive/distributional MDPs.
% The concept of optimistic or risk-averse \etc should also be understood this way.
% It is biased.
% How does the path ensemble estimation different from the real estimation.
% But it can approximately achieve similar effects via pragmatical way.
%
With this in mind,
we discuss on the effects of these biased estimations in the following.

\paragraph{The \(\max\) and the \(\min\) Statistics}
The \(\max\) statistics leads to optimistic estimation of advantages.
% This means that for all possible trajectories following the policy,
% it may increase the advantage of bad paths.
% Then actions that may lead to good states will not be buried by the later bad actions.
% This is beneficial for sparse environments.
It is beneficial for sparse-reward environments as actions that may lead to large returns will not be buried by the later bad actions.
%, as we will verify in experiments.
The \(\min\) statistics implies the risk-averse bias.
It avoids actions that may cause bad states later.
By decreasing the advantage estimation of those actions,
it makes the optimization direction away from them.
% This risk-averse is explained as,
% along any trajectory, we avoid bad actions happens.
This property is useful for fragile environments,
such as the biped locomotion environments which are sensitive to joint motions of the characters.

\paragraph{The \(\maxabs\) Statistics}
In the \(\maxabs\) statistics,
the estimator with the largest absolute value in the path ensemble is chosen.
This means that an action is evaluated as either overly good when the largest positive advantage is selected,
or overly bad when the smallest negative advantage is selected.
We refer to it as an exaggerated estimation.
% It partitions the estimations in the path ensemble into two groups,
% where the positive estimations belong to one group and the remaining belong to the other.
% It compares which group is more strong, which is measured by the maximum absolute value of the set.
% Then the winner partition assigns its estimation value with the largest absolute value.
This statistics implements a heuristic that makes the good paths look better and the bad paths look worse.
The bias generally improves sample efficiency for the MuJoCo environments and the Atari environments,
as will be shown in the experiment section.
% \(\min\circ\abs\) statistics, on the other hand, selects the value with the minimum absolute value.
% It is a way to reduce variance, thus stabilizes the training.
% Sometimes it performs very well for some special environments.
% which means that it decide whether the action should be encouraged or discouraged based on the the relative ``strength'' of two parties.
% Application to continuous control, sample efficiency

% Low variance.

\paragraph{The General Order Statistics}
The values of the general order statistics lie between the maximum and the minimum.
% To our surprise,
% in many environments,
But the performance of general order statistics is not simply an interpolation between the performance of the \(\max\) statistics and the \(\min\) statistics, as shown in Figure~\ref{fig:mujoco_rank}.
% Selecting the estimator that was in a prescribed order lies between the above discussed extreme cases.
In many environments, the intermediate order statistics performs worse than the two extreme cases,
% rather than having a in-between performance.
which indicates that the \(\max\) statistics and the \(\min\) statistics indeed have special explanations,
such as the optimistic viewpoint and the risk-averse viewpoint we have discussed.
Therefore, that extreme cases of \(\max\), \(\min\) and \(\maxabs\) might have more applications than the general order statistics.

\section{Experiment}\label{sec:exp}

We first evaluate the performance of our algorithm on four different types of problems,
and then we study the effects of the hyper-parameters.

\subsection{Sparse Reward and the Optimistic Exploration}

\begin{figure}[t]
    \centering
    \includegraphics[width=0.8\linewidth]{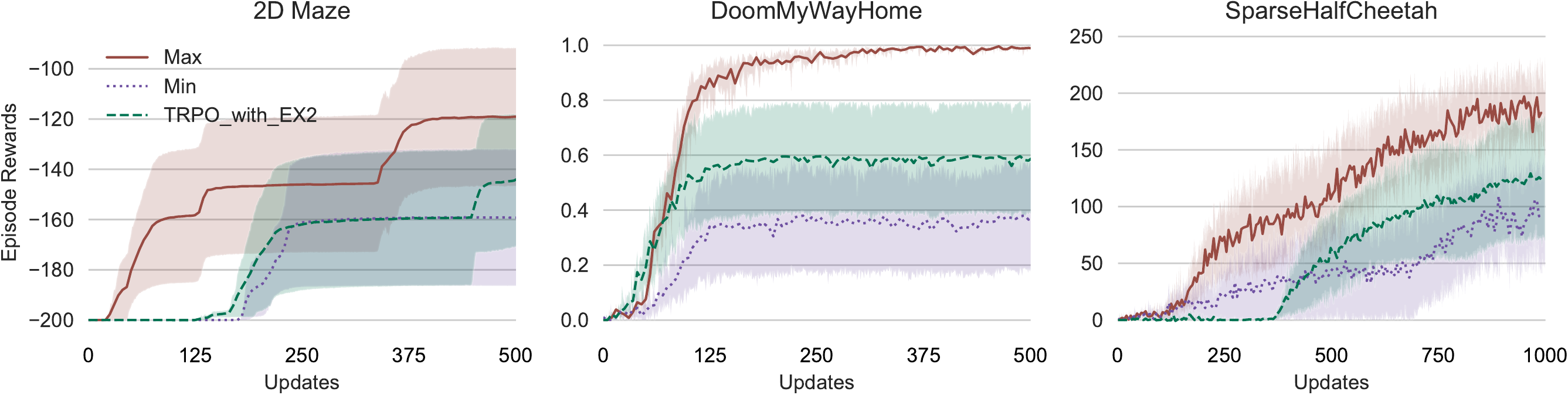}
	\caption{
        The \(\max\) and the \(\min\) statistics on three different types of sparse-reward environments.
        The \(\max\) statistics were significantly better than the original EX2 algorithm on all three environments.
        The \(\min\) statistics were not effective for sparse reward tasks.
        All settings were run for \(5\) seeds.
	}\label{fig:sparse_max}
\end{figure}

The amortized EX2 algorithm~\cite{NIPS2017_6851} is specially designed for tasks with sparse rewards.
Three environments were chosen from their paper, which represent three types of problems.
% They are the 2D Maze for low-dimensional tasks,
% the SparseHalfCheetah for medium-dimensional continuous control
% and the Doom game for high-dimensional image-based control.
The biased estimation over path ensemble was implemented based on the original code from the paper,
where the TRPO is used for policy optimization.
We modified their advantage estimation procedure,
while left other parts untouched.
The path ensemble was composed of \(k\)-step estimators where \(k\in\{1, 16, 64, 4000\}\) for Maze and Doom,
and \(k\in\{1, 16, 64, 5000\}\) for SparseHalfCheetah.
The numbers \(4000\) and \(5000\) originated from the respective batch sizes.
The bias ratio for both the \(\max\) and the \(\min\) statistics was set to \(\rho=0.5\) for Maze and Doom,
and \(\rho=0.3\) for SparseHalfCheetah.
All hyper-parameters followed the settings of the amortized EX2.
The result is shown in Figure~\ref{fig:sparse_max}.

\paragraph{2D Maze}
This environment provides a sparse reward function,
where the agent can only get the reward when it is within a small radius of the goal.
From the figure, we observed that the \(\max\) statistics started to gain rewards at the very early stage,
which indicates the optimistic bias provides a supreme sample efficiency over other methods.
After \(500\) updates, the average episodic reward of \(\max\) was much higher than that of the EX2 algorithm.
The \(\min\) statistics was worse than EX2.

\paragraph{Doom}
MyWayHome is a vision-based maze navigation benchmark,
where the agent is required to navigate through a series of interconnected rooms before reaching the goal.
A \(+1\) reward is given for reaching the goal before timeout.
The task is challenging because the inputs are realistic images in the first-person perspective.
The \(\max\) statistics was very effective for this challenge and it reached nearly \(100\%\) success rate.
In comparison, EX2 only got an average of \(0.6\) episode rewards.

\paragraph{SparseHalfCheetah}
It is a challenging continuous control task with sparse reward.
Our optimistic exploration significantly improved the sample efficiency,
and the reward after \(1000\) updates was \(40\%\) higher than the EX2 algorithm.
Note that the \(\min\) statistics was less effective than EX2 in all three benchmarks,
which demonstrates that risk-averse strategies are not effective in sparse environments.

\subsection{Locomotion and the Risk-Averse Exploration}\label{sec:loco}

\begin{figure}[t]
	\centering
	\includegraphics[width=0.15\linewidth,valign=m]{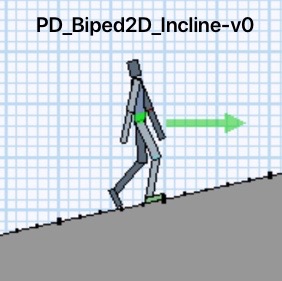}\hspace{0.3cm}
	\includegraphics[width=0.15\linewidth,valign=m]{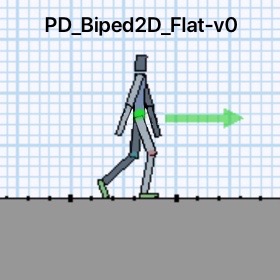}\hspace{0.5cm}
	\includegraphics[width=0.5\linewidth,valign=m]{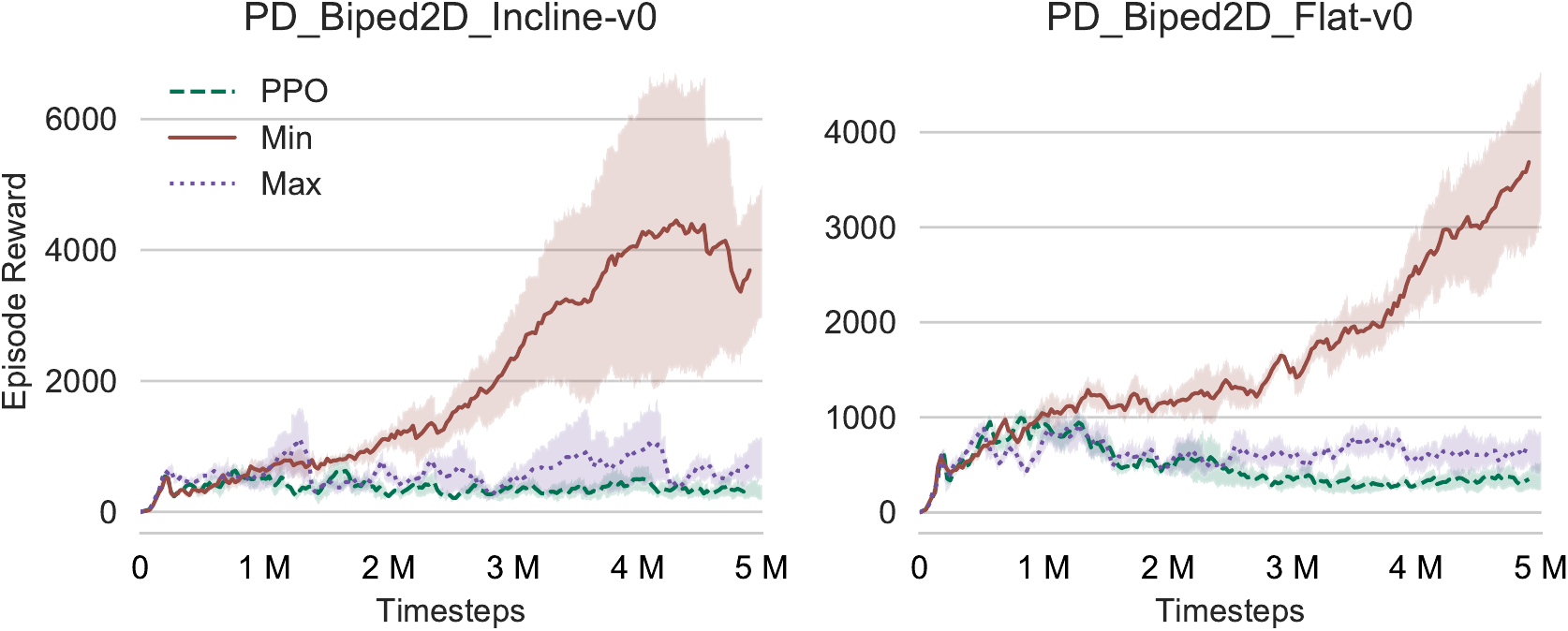}
	\caption{
		Left: screenshot of the flat and incline environment for biped walking.
		% Ablation study on order of the path ensemble on Terrain Biped Walking environment.
		Right: the performance of PPO and PPO with \(\min\) and \(\max\) statistics.
		% The baseline PPO and the \(\max\) cannot master the tasks.
		% The \(\min\) with risk-averse exploration is preferred.
		% So Biped Walking prefers the conservative exploration.
		We ran 5 seeds for each setting.
	}\label{fig:terrain}
\end{figure}

The \(\min\) statistics is very useful for environments that are fragile to actions,
\ie, one wrong action would lead to catastrophic results later on.
The Terrain RL simulator~\cite{Berseth2018TerrainRS} provides such environments.
Two biped walking tasks were selected, where a character needs to learn how to walk steadily forward in two different terrains,
the flat and the incline, as shown in Figure~\ref{fig:terrain}.
The character continuously receives rewards until it fells down or is stuck, in which case the episode terminates.
This task is challenging because the observation only contains the pose of the character,
which forces the character to learn how to walk without memorizing actions based on its location.
In contrast, the locomotion tasks in MuJoCo environments contain absolute world coordinates in the observation.
The action space is \(11\)-dimensional, which corresponds to the joints of the character.
The environment is fragile.
If a wrong action is performed,
the character might lose balance and then fall down.
We used PPO in this experiment.
The hyper-parameters were borrowed from those designed for MuJoCo environments in the baselines'~\cite{baselines} PPO implementation.
The \(\min\) and \(\max\) statistics were implemented on top of the PPO algorithm.
The path ensemble consisted of \(k\)-step estimators where \(k\in\{1, 16, 64, 2048\}\),
% where \(2048\) is the batch size.
and the bias ratio is \(\rho=0.3\).
The results are shown in Figure~\ref{fig:terrain}.
The risk-averse policy via the \(\min\) statistics successfully mastered the task while the vanilla PPO algorithm failed.
The optimistic strategy by the \(\max\) was not very effective in these environments
as the low reward curves indicate that the character fells down at the beginning of the episode.
We conclude that the risk-averse exploration by the \(\min\) statistics of the path ensemble helps the agent to learn in fragile environments.

\subsection{Continuous Control and the Exaggerated Exploration}

\begin{figure}[t]
	\centering
	\includegraphics[width=0.99\linewidth]{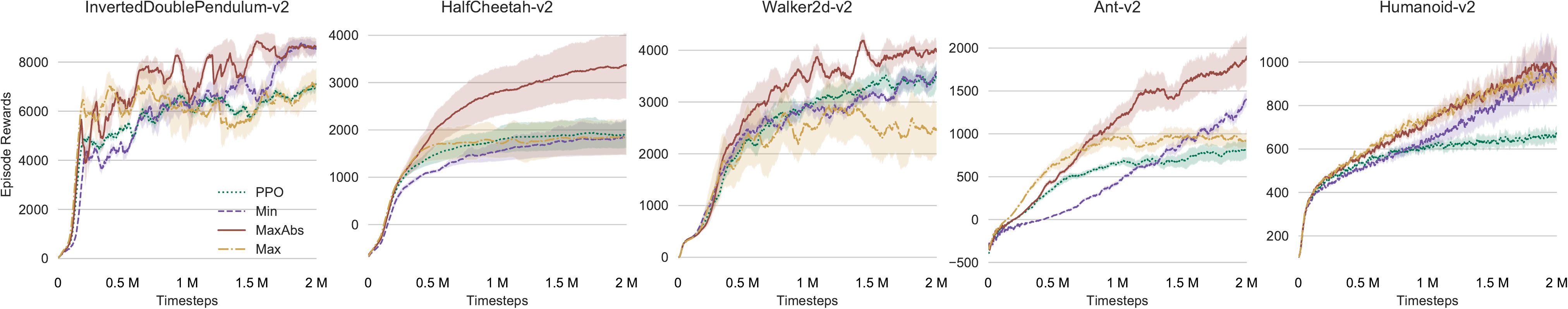}
	\caption{
		Experiments on MuJoCo environments.
		Three biased estimates \(\max\), \(\min\) and \(\maxabs\) were compared with the vanilla PPO algorithm.
		% The exaggerated exploration strategy by the \(\maxabs\) statistics has better performance than others.
		Each setting was run with \(5\) seeds.
	}\label{fig:mujoco_abs}
\end{figure}

We tested various biased advantage estimations on \(5\) continuous control benchmarks based on the MuJoCo physics simulator.
They are not sparse-reward environments,
and most of them are also not sensitive to individual actions.
For example, the design of the HalfCheetah agent makes it hard to fall down or be stuck.
% environment is a 2D robot run task,
% in which the  .
We used PPO in this set of experiments.
The implementation was based on the baselines~\cite{baselines} code repository,
and the default hyper-parameters were adopted.
We tested three biased estimations including the \(\max\), the \(\min\) and the \(\maxabs\) statistics.
In all settings, the path ensemble had index set \(\{1, 16, 64, 2048\}\),
and the bias ratio was set to \(\rho=0.4\).
Results are shown in Figure~\ref{fig:mujoco_abs}.
We observed that the \(\maxabs\) statistics was consistently better than the baseline PPO algorithm,
while the performances of the \(\max\) and the \(\min\) statistics depended upon the specific environment.
So we conclude that the exaggerated exploration by the \(\maxabs\) statistics is generally effective for a wide range of environments.
% if no special properties such as sparsity present.

\subsection{Atari Games}

\begin{figure}[t]
    \centering
    \includegraphics[width=1\linewidth]{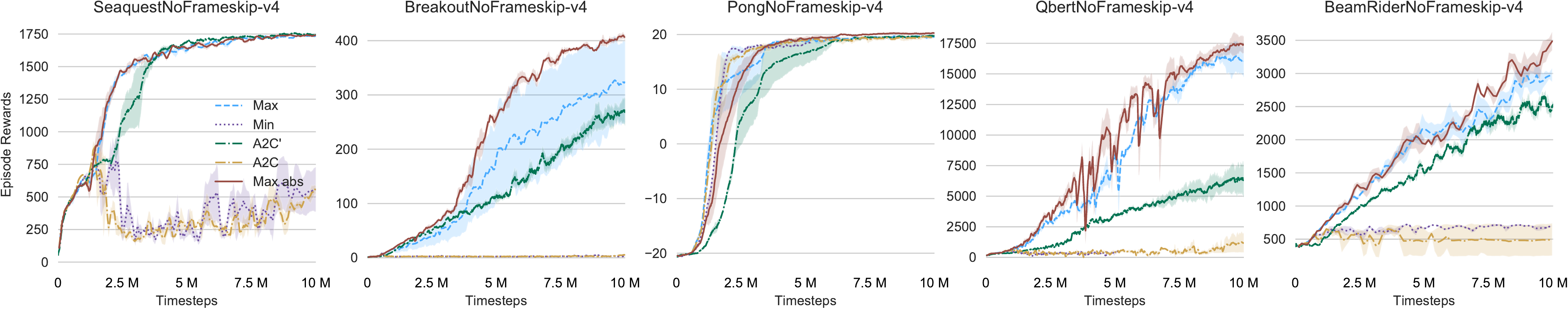}
    \caption{
		Experiments on the Atari environments.
		A2C' was the original A2C algorithm without mini-batch,
		and A2C was using the same mini-batch configuration as other three settings: \(\max\), \(\min\) and \(\maxabs\).
        % The exaggerated strategy by the \(\maxabs\) statistics performs generally well on all these environments.
		Each setting was run for \(3\) seeds.
		(This figure is best viewed in color.)
}\label{fig:atari}
\end{figure}

The biased advantage estimations were also tested on a subset of Atari game environments.
We used the A2C algorithm with \(4\) paralleled threads.
The implementation was based on the A2C codes in baselines~\cite{baselines}.
Hyper-parameters were the defaults for Atari games.
Since the sampled trajectory length was \(5\) in the default setting,
the path ensemble had fewer elements, which was composed of \(\{1, 3, 5\}\).
The maximum trajectory length of \(5\) in a batch was too small to gather enough data on \(k\)-step estimations when \(k>1\).
For example, only the first state has a valid \(5\)-step advantage estimation in a length-\(5\) trajectory.
This affects the power of the path ensemble.
To circumvent this limitation,
we collected paths of length \(5n\),
and then computed the biased advantage estimators using these longer trajectories.
Since the batch size was \(n\) times of the original setting,
it was split into \(n\) mini-batches.
For Seaquest and Breakout, \(n=20\);
for Pong and Qbert, \(n=10\);
and for BeamRider, \(n=60\).
The result is shown in Figure~\ref{fig:atari}.
In these environments,
the sparsity and fragility were unknown.
Obviously, the exaggerated exploration generally improves the performance,
whereas the \(\max\) and the \(\min\) statistics
are only effective for a subset of environments.
% This is similar to the results on MuJoCo environments.

\subsection{Ablation Study}

\begin{figure}[t]
	\centering
	\includegraphics[width=1\linewidth]{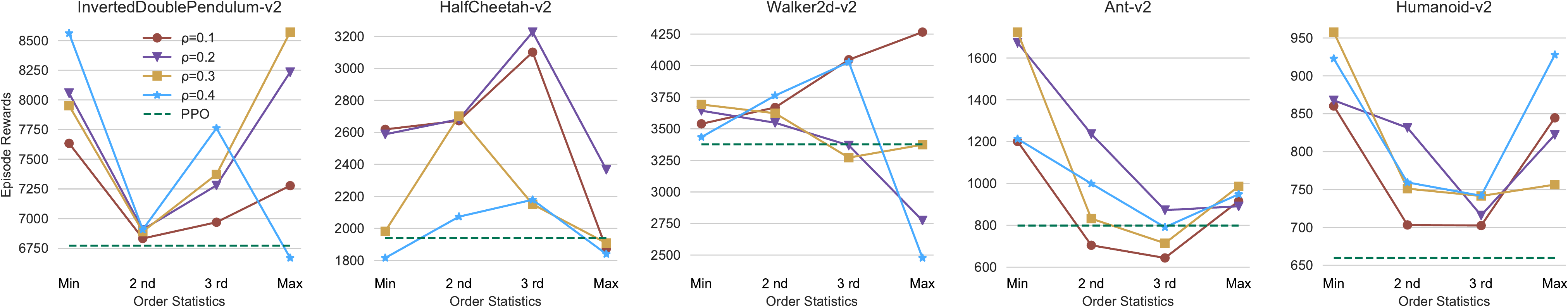}
	\caption{
		The order statistics over the path ensemble in MuJoCo environments.
		\(x\)-axis labels the different order statistics.
		The four series in the sub-figures were varying the bias ratio \(\rho\),
		and the flat line indicated performance of the baseline PPO algorithm.
		% When \(\rho\) is too large, the performance may degrade (\eg in HalfCheetah).
		Each data point was averaged over \(5\) runs.
		% Since there are 4 estimators in the path ensemble, 0.25 is the minimum and 1.0 is the maximum in the path ensemble.
		% Different series correspond to hyper-parameter of choosing probability.
		% Different environments have different performance profiles.
		% Humanoid and InvertedDoublePendulum welcomes either min or max,
		% Ant and InvertedPendulum prefers pessimistic estimations.
		% Hopper and HalfCheetah prefers optimistic estimation but a little sensitive to extreme estimations, so the second largest estimation is best.
	}\label{fig:mujoco_rank}
\end{figure}

In this section we study the behavior of general order statistics on MuJoCo environments,
and then discuss the sensitivity of the algorithms to hyper-parameters.
%  of bias selection rate \(\rho\).

\paragraph{The General Order Statistics}
As shown in Figure~\ref{fig:mujoco_rank},
in three (InvertedDoublePendulum,  Ant, and Humanoid) of the five environments,
the \(\max\) and \(\min\) were better than the \(2\)nd and the \(3\)rd order statistics.
This means that some environments welcome both optimistic and risk-averse exploration.
Overall,
the intermediate order statistics was better than the baseline,
which used the GAE estimator.

\paragraph{Sensitivity on the Bias Ratio}
Generally speaking, the bias ratio \(\rho\) had a great influence on the performance.
As shown in Figure~\ref{fig:mujoco_rank},
the range \([0.2, 0.4]\) is a plausible choice for most experiments.
If the ratio is further increased, the performance may degrade.

\paragraph{Effect of Ensemble Size}

We built another path ensemble whose index set consisted of \(12\) elements.
They were \(\{1, 2, 4, 8, 16, 32, 64, 128, 256, 512, 1024, 2048\}\).
Final performances of the three order statistics, \(\max\), \(\min\) and \(\maxabs\),
were tested under the bias ratio \(\rho=0.4\).
For the ensemble with \(4\) elements \(\{1, 16, 64, 2048\}\),
the average episode reward in the end of the training was \(928, 927, 974\), respectively;
and for the ensemble of size \(12\), the numbers were \(767, 854\) and \(867\).
It shows that a larger ensemble does not necessarily lead to better performance.

\section{Conclusion}\label{sec:conclusion}
In this paper, we proposed a simple yet effective way of exploration via the biased estimate of advantages.
They were implemented through the family of order statistics over the path ensembles,
which formed nonlinear combinations of different \(k\)-step estimators.
The maximum, the minimum, and the element with the maximum absolute value were studied in detail.
We incorporated these biased advantage estimators into three widely used actor-critic algorithms including A2C, TRPO and PPO\@.
With different biased estimations,
the proposed algorithm could be effective in solving sparse-reward environments,
or the fragile environments which are sensitive to individual actions.
We verified the effectiveness of our approach by extensive experiments on various domains,
including the continuous control, locomotion, video games, and sparse-reward environments.

% \clearpage

{\small
  \bibliographystyle{ieee}
  \bibliography{main}

\begin{thebibliography}{10}\itemsep=-1pt

\bibitem{barth2018distributed}
G.~Barth-Maron, M.~W. Hoffman, D.~Budden, W.~Dabney, D.~Horgan, A.~Muldal,
  N.~Heess, and T.~Lillicrap.
\newblock Distributed distributional deterministic policy gradients.
\newblock {\em arXiv preprint arXiv:1804.08617}, 2018.

\bibitem{bellemare2017distributional}
M.~G. Bellemare, W.~Dabney, and R.~Munos.
\newblock A distributional perspective on reinforcement learning.
\newblock In {\em Proceedings of the 34th International Conference on Machine
  Learning-Volume 70}, pages 449--458. JMLR. org, 2017.

\bibitem{bellemare13arcade}
M.~G. {Bellemare}, Y.~{Naddaf}, J.~{Veness}, and M.~{Bowling}.
\newblock The arcade learning environment: An evaluation platform for general
  agents.
\newblock {\em Journal of Artificial Intelligence Research}, 47:253--279, jun
  2013.

\bibitem{Berseth2018TerrainRS}
G.~Berseth, X.~B. Peng, and M.~van~de Panne.
\newblock Terrain rl simulator.
\newblock {\em CoRR}, abs/1804.06424, 2018.

\bibitem{gym}
G.~Brockman, V.~Cheung, L.~Pettersson, J.~Schneider, J.~Schulman, J.~Tang, and
  W.~Zaremba.
\newblock Openai gym, 2016.

\bibitem{chow2014algorithms}
Y.~Chow and M.~Ghavamzadeh.
\newblock Algorithms for cvar optimization in mdps.
\newblock In {\em Advances in neural information processing systems}, pages
  3509--3517, 2014.

\bibitem{chow2015risk}
Y.~Chow, A.~Tamar, S.~Mannor, and M.~Pavone.
\newblock Risk-sensitive and robust decision-making: a cvar optimization
  approach.
\newblock In {\em Advances in Neural Information Processing Systems}, pages
  1522--1530, 2015.

\bibitem{dabney2018implicit}
W.~Dabney, G.~Ostrovski, D.~Silver, and R.~Munos.
\newblock Implicit quantile networks for distributional reinforcement learning.
\newblock In {\em International Conference on Machine Learning}, pages
  1104--1113, 2018.

\bibitem{delage2007percentile}
E.~Delage and S.~Mannor.
\newblock Percentile optimization in uncertain markov decision processes with
  application to efficient exploration.
\newblock In {\em Proceedings of the 24th international conference on Machine
  learning}, pages 225--232. ACM, 2007.

\bibitem{delage2010percentile}
E.~Delage and S.~Mannor.
\newblock Percentile optimization for markov decision processes with parameter
  uncertainty.
\newblock {\em Operations research}, 58(1):203--213, 2010.

\bibitem{baselines}
P.~Dhariwal, C.~Hesse, O.~Klimov, A.~Nichol, M.~Plappert, A.~Radford,
  J.~Schulman, S.~Sidor, Y.~Wu, and P.~Zhokhov.
\newblock Openai baselines.
\newblock \url{https://github.com/openai/baselines}, 2017.

\bibitem{NIPS2017_6851}
J.~Fu, J.~Co-Reyes, and S.~Levine.
\newblock Ex2: Exploration with exemplar models for deep reinforcement
  learning.
\newblock In I.~Guyon, U.~V. Luxburg, S.~Bengio, H.~Wallach, R.~Fergus,
  S.~Vishwanathan, and R.~Garnett, editors, {\em Advances in Neural Information
  Processing Systems 30}, pages 2577--2587. Curran Associates, Inc., 2017.

\bibitem{fujimoto2018addressing}
S.~Fujimoto, H.~van Hoof, and D.~Meger.
\newblock Addressing function approximation error in actor-critic methods.
\newblock {\em arXiv preprint arXiv:1802.09477}, 2018.

\bibitem{heess2017emergence}
N.~Heess, S.~Sriram, J.~Lemmon, J.~Merel, G.~Wayne, Y.~Tassa, T.~Erez, Z.~Wang,
  S.~Eslami, M.~Riedmiller, et~al.
\newblock Emergence of locomotion behaviours in rich environments.
\newblock {\em arXiv preprint arXiv:1707.02286}, 2017.

\bibitem{howard1972risk}
R.~A. Howard and J.~E. Matheson.
\newblock Risk-sensitive markov decision processes.
\newblock {\em Management science}, 18(7):356--369, 1972.

\bibitem{mannor2012lightning}
S.~Mannor, O.~Mebel, and H.~Xu.
\newblock Lightning does not strike twice: robust mdps with coupled
  uncertainty.
\newblock In {\em Proceedings of the 29th International Coference on
  International Conference on Machine Learning}, pages 451--458. Omnipress,
  2012.

\bibitem{mnih2016asynchronous}
V.~Mnih, A.~P. Badia, M.~Mirza, A.~Graves, T.~Lillicrap, T.~Harley, D.~Silver,
  and K.~Kavukcuoglu.
\newblock Asynchronous methods for deep reinforcement learning.
\newblock In {\em International conference on machine learning}, pages
  1928--1937, 2016.

\bibitem{morimura2010nonparametric}
T.~Morimura, M.~Sugiyama, H.~Kashima, H.~Hachiya, and T.~Tanaka.
\newblock Nonparametric return distribution approximation for reinforcement
  learning.
\newblock In {\em Proceedings of the 27th International Conference on Machine
  Learning (ICML-10)}, pages 799--806, 2010.

\bibitem{nilim2004robustness}
A.~Nilim and L.~El~Ghaoui.
\newblock Robustness in markov decision problems with uncertain transition
  matrices.
\newblock In {\em Advances in Neural Information Processing Systems}, pages
  839--846, 2004.

\bibitem{prashanth2014policy}
L.~Prashanth.
\newblock Policy gradients for cvar-constrained mdps.
\newblock In {\em International Conference on Algorithmic Learning Theory},
  pages 155--169. Springer, 2014.

\bibitem{prashanth2013actor}
L.~Prashanth and M.~Ghavamzadeh.
\newblock Actor-critic algorithms for risk-sensitive mdps.
\newblock In {\em Advances in neural information processing systems}, pages
  252--260, 2013.

\bibitem{schulman2015trust}
J.~Schulman, S.~Levine, P.~Abbeel, M.~Jordan, and P.~Moritz.
\newblock Trust region policy optimization.
\newblock In {\em International Conference on Machine Learning}, pages
  1889--1897, 2015.

\bibitem{schulman2015high}
J.~Schulman, P.~Moritz, S.~Levine, M.~Jordan, and P.~Abbeel.
\newblock High-dimensional continuous control using generalized advantage
  estimation.
\newblock {\em arXiv preprint arXiv:1506.02438}, 2015.

\bibitem{schulman2017proximal}
J.~Schulman, F.~Wolski, P.~Dhariwal, A.~Radford, and O.~Klimov.
\newblock Proximal policy optimization algorithms.
\newblock {\em arXiv preprint arXiv:1707.06347}, 2017.

\bibitem{seijen2014true}
H.~Seijen and R.~Sutton.
\newblock True online td (lambda).
\newblock In {\em International Conference on Machine Learning}, pages
  692--700, 2014.

\bibitem{smirnova2019distributionally}
E.~Smirnova, E.~Dohmatob, and J.~Mary.
\newblock Distributionally robust reinforcement learning.
\newblock {\em arXiv preprint arXiv:1902.08708}, 2019.

\bibitem{sutton2011reinforcement}
R.~S. Sutton and A.~G. Barto.
\newblock Reinforcement learning: An introduction.
\newblock 2011.

\bibitem{tamar2012policy}
A.~Tamar, D.~Di~Castro, and S.~Mannor.
\newblock Policy gradients with variance related risk criteria.
\newblock In {\em Proceedings of the 29th International Coference on
  International Conference on Machine Learning}, pages 1651--1658. Omnipress,
  2012.

\bibitem{tamar2015optimizing}
A.~Tamar, Y.~Glassner, and S.~Mannor.
\newblock Optimizing the cvar via sampling.
\newblock In {\em Twenty-Ninth AAAI Conference on Artificial Intelligence},
  2015.

\bibitem{vanHasselt2012}
H.~van Hasselt.
\newblock {\em Reinforcement Learning in Continuous State and Action Spaces},
  pages 207--251.
\newblock Springer Berlin Heidelberg, Berlin, Heidelberg, 2012.

\bibitem{van2007reinforcement}
H.~Van~Hasselt and M.~A. Wiering.
\newblock Reinforcement learning in continuous action spaces.
\newblock In {\em 2007 IEEE International Symposium on Approximate Dynamic
  Programming and Reinforcement Learning}, pages 272--279. IEEE, 2007.

\bibitem{vinyals2017starcraft}
O.~Vinyals, T.~Ewalds, S.~Bartunov, P.~Georgiev, A.~S. Vezhnevets, M.~Yeo,
  A.~Makhzani, H.~K{\"u}ttler, J.~Agapiou, J.~Schrittwieser, et~al.
\newblock Starcraft ii: A new challenge for reinforcement learning.
\newblock {\em arXiv preprint arXiv:1708.04782}, 2017.

\bibitem{watkins1989learning}
C.~J. C.~H. Watkins.
\newblock Learning from delayed rewards.
\newblock 1989.

\bibitem{xu2007robustness}
H.~Xu and S.~Mannor.
\newblock The robustness-performance tradeoff in markov decision processes.
\newblock In {\em Advances in Neural Information Processing Systems}, pages
  1537--1544, 2007.

\end{thebibliography}
}

\end{document}